\documentclass[conference]{IEEEtran}
\usepackage[utf8]{inputenc}
\usepackage{cite}
\usepackage{amsmath,amssymb,amsfonts}
\usepackage{algorithmic}
\usepackage{graphicx}
\usepackage{textcomp}
\usepackage{xcolor}
\usepackage{array}
\usepackage[caption=false,font=normalsize,labelfont=sf,textfont=sf]{subfig}
\usepackage{textcomp}
\usepackage{stfloats}
\usepackage{url}
\usepackage{verbatim}
\usepackage{graphicx}
\usepackage{bm}
\usepackage{cite}
\usepackage{times}
\usepackage{epsfig}
\usepackage{graphicx}
\usepackage{adjustbox}
\usepackage{tabularx} 
\usepackage{utfsym}
\usepackage{amsmath}
\usepackage{amssymb}
\usepackage{multirow}
\usepackage{booktabs}
\usepackage{xcolor}
\usepackage{lipsum}
\usepackage{subcaption} 
\usepackage{multirow}
\usepackage{makecell} 
\usepackage[linesnumbered, ruled]{algorithm2e}
\usepackage[table]{xcolor} 
\usepackage{hyperref}
\usepackage{enumitem}
\usepackage[normalem]{ulem}

\def\BibTeX{{\rm B\kern-.05em{\sc i\kern-.025em b}\kern-.08em
    T\kern-.1667em\lower.7ex\hbox{E}\kern-.125emX}}
\begin{document}

\title{PoseStreamer: A Multi-modal Framework for 3D Tracking of Unseen Moving Objects}

\author{
    \large 
    Huiming Yang\textsuperscript{1}, 
    Linglin Liao\textsuperscript{1}, 
    Fei Ding, 
    Sibo Wang, 
    Zijian Zeng\textsuperscript{$\dagger$} \\[0.5em] 
    
    \normalsize 
    School of Mathematics, Renmin University of China, Beijing, China \\[0.5em] 
    
    \ttfamily \small 
    \{xuihaipiaoxiang0, liaolinglin6, shawnzengquant\}@gmail.com, \\
    dingfei@email.ncu.edu.cn, ai\_sibo@sina.com
}

\maketitle

\renewcommand{\thefootnote}{\fnsymbol{footnote}}
\footnotetext[2]{Corresponding author.}

\begin{abstract}
Six degree of freedom (6DoF) pose estimation for unseen objects is a critical task in computer vision, yet it faces significant challenges in high-speed and low-light scenarios where standard RGB cameras suffer from motion blur. While event cameras offer a promising solution due to their high temporal resolution, current 6DoF pose estimation methods typically yield suboptimal performance in high-speed object moving scenarios. To address this gap, we propose \textbf{PoseStreamer}, a robust multi-modal 6DoF pose estimation framework designed specifically on high-speed moving scenarios. Our approach integrates three core components: an \textbf{Adaptive Pose Memory Queue} that utilizes historical orientation cues for temporal consistency, an \textbf{Object-centric 2D Tracker} that provides strong 2D priors to boost 3D center recall, and a \textbf{Ray Pose Filter} for geometric refinement along camera rays. Furthermore, we introduce \textbf{MoCapCube6D}, a novel multi-modal dataset constructed to benchmark performance under rapid motion. Extensive experiments demonstrate that PoseStreamer not only achieves superior accuracy in high-speed moving scenarios, but also exhibits strong generalizability as a template-free framework for unseen moving objects.
\end{abstract}

\begin{IEEEkeywords}
6DoF Pose Estimation, Multi-modal Fusion, High-speed Tracking, Unseen Objects, Event-based Vision
\end{IEEEkeywords}

\section{Introduction}
\label{sec:intro}

6DoF pose estimation\cite{pavlakos20176} aims to compute the rigid six-dimensional transformation (3D translation and rotation) between an object and a camera. Traditional 6DoF pose estimation methods\cite{sundermeyer2018implicit,labbe2020cosypose,xiang2018posecnn} cannot be directly applied to unseen objects during the training phase. In practical scenarios, different applications often provide diverse input objects. More recent efforts have focused on real-time pose estimation for arbitrary unseen objects. However, existing RGB camera-based methods\cite{wen2024foundationpose} are prone to motion blur, which particularly limits their performance in high-speed low-lit scenarios.

\begin{figure*}[htbp]
    \centering
    \includegraphics[width=1.0\textwidth]{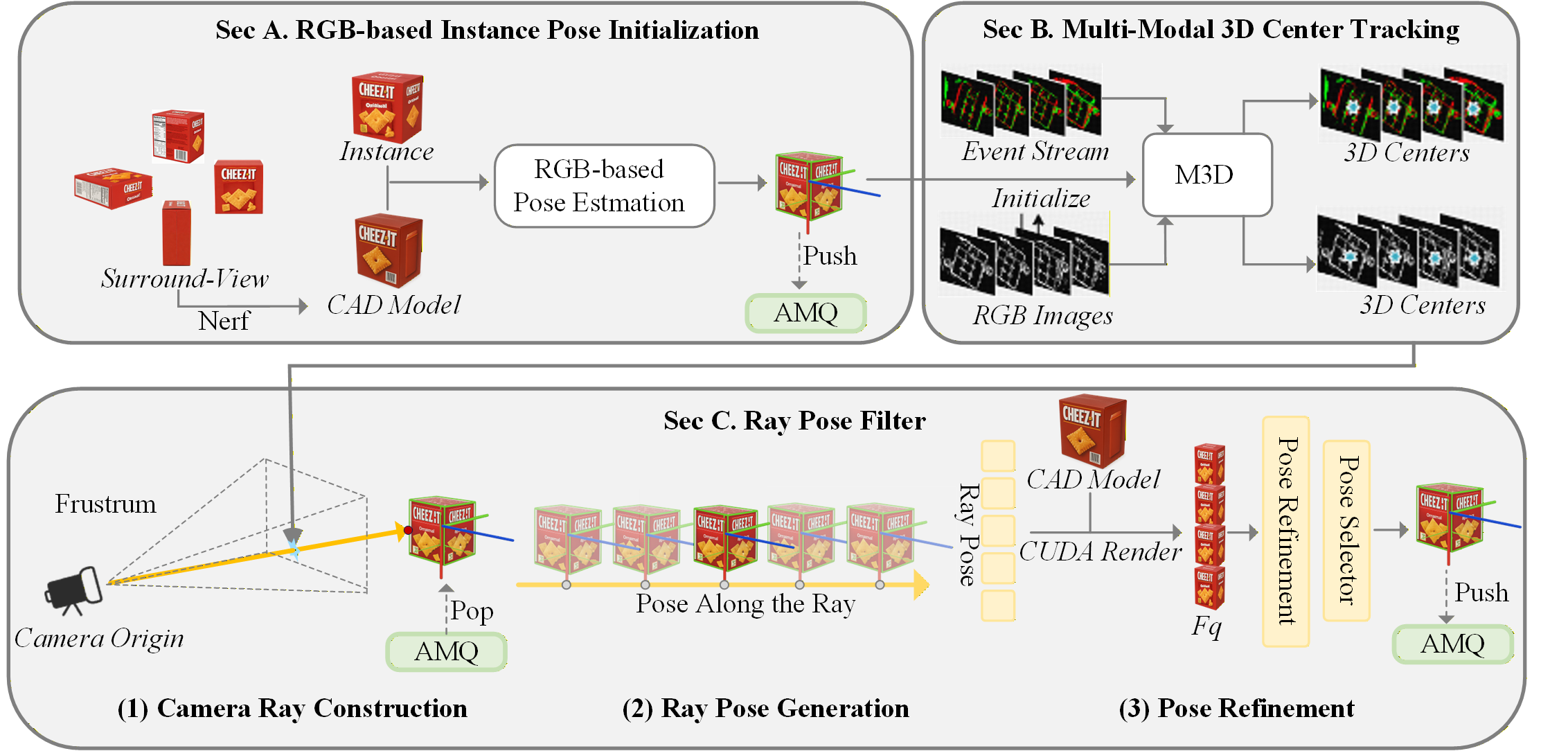} 
    \caption{Overview of the PoseStreamer Architecture. The framework proceeds in three stages: (A) Unseen object initialization via RGB-based reconstruction and the Adaptive Pose Memory Queue (AMQ). (B) High-speed 3D center estimation via the Multi-modality 3D Tracker (M3D) on RGB images and Event streams. (C) Fine-grained 6DoF optimization via the Ray Pose Filter (RPF). The filter samples and selects pose hypotheses along the camera ray.}
    \label{fig:main}
\end{figure*}

Some studies introduce event cameras\cite{gallego2020event} which capture motion information via luminance variations. It outputs a sparse event stream with high temporal resolution. Such a stream can act as a complementary modality to the motion geometry information for RGB cameras. Currently, relevant research on visual object tracking using event cameras remains relatively limited. EventVOT \cite{wang2024event} distills knowledge from an RGB-Event dual-modal teacher network into a pure event-driven student network. AFNet \cite{zhang2023frame} fuses frame-based RGB data and event-based temporal data through a multi-modal alignment and fusion module. STNet \cite{zhang2022spiking} integrates Transformer with spiking neural networks to build an event-driven tracking framework. However, all the aforementioned methods only validate 2D tracking performance and do not explore object tracking in the 3D space. Our research aims to leverage the RGB-Event fused modality to achieve accurate 6DoF pose estimation for arbitrary moving objects. On the one hand, event cameras only capture edge information from luminance variations. They lack critical contextual details, such as texture, color, and global semantics, which are essential for object orientation estimation. Moreover, a significant distribution gap exists between event data and RGB modality data\cite{wang2023visevent, tang2025revisiting}, posing a barrier to direct fusion for information complementarity.
On the other hand, event cameras output relatively sparse spatial signals. This sparsity can lead to tracking failure when target object contours are insufficiently distinct.\cite{messikommer2025data} In addition, the temporal resolution mismatch between event and RGB cameras further amplifies the uncertainty of spatial perception.

To address the aforementioned challenges, this paper proposes \textbf{PoseStreamer}, a multi-modal 6DoF pose estimation framework tailored for high-speed moving scenarios. Its core architecture comprises three key components: (1) The \uline{A}daptive Pose \uline{M}emory \uline{Q}ueue \textbf{(AMQ)}, a sliding-window module that leverages historical orientation cues to guide current-frame orientation inference. (2) The \uline{M}ulti-modality \uline{3D} tracker \textbf{(M3D)}, which supplies robust 2D priors to enhance the recall of 3D object centers. (3) The \uline{R}ay \uline{P}ose \uline{F}ilter \textbf{(RPF)}, designed to refine pose estimates along camera ray directions. Furthermore, to evaluate performance under high-speed motion, we introduce \textbf{MoCapCube6D}, a novel multi-modal pose estimation dataset that facilitates comprehensive benchmarking of our method against state-of-the-art approaches. Notably, the proposed method exhibits strong generalizability and can be extended to a template-free framework for unseen object instances. Our contributions are summarized as follows:
\begin{itemize}
    \item We propose \textbf{PoseStreamer}, a novel multi-modal 6DoF pose estimation framework. By integrating the Adaptive Pose Memory Queue, Multi-modality 3D tracker, and Ray Pose Filter, our method effectively leverages historical cues and 2D priors to achieve robust tracking on high-speed moving scenarios.
    \item We demonstrate the strong generalizability of our approach by extending it to a \textbf{template-free framework} capable of handling arbitrary unseen object instances without retraining, addressing the limitations of traditional methods on unseen objects.
    \item We construct \textbf{MoCapCube6D}, a multi-modal benchmark dataset for 6DoF pose estimation performance on high-speed moving scenarios.
\end{itemize}
\section{Related Works}
\label{sec:related}

\subsection{6DoF Pose Estimation and Tracking}
6DoF pose estimation aims to infer the 3D translation and rotation of a target object. Traditional methods often require instance- or category-level CAD models for offline training or template matching~\cite{sundermeyer2018implicit}, which restricts their application to unseen objects. Although recent generalizable works~\cite{liu2022gen6d} relax these assumptions, they typically rely on pre-captured reference views of the test object. In the tracking domain, methods like BundleTrack~\cite{wen2021bundletrack} attempt to generalize to unseen objects instantly without prior templates. However, these RGB-based approaches rely heavily on clear textures and accurate feature matching. Consequently, they suffer significant performance degradation in high-speed scenarios where severe motion blur\cite{rozumnyi2017fastmoving} and artifacts sever the required 2D-3D correspondences.

\subsection{Event Camera-based Detection and Tracking}
Event cameras excel in high-speed scenarios due to their microsecond-level latency. Early works like EKLT~\cite{gehrig2020eklt} fused frames and events to track visual features asynchronously. Recent data-driven methods focus on enhancing modality interaction; for instance, Wang et al.~\cite{wang2023visevent} utilize cross-modality Transformers to fuse RGB and event streams, while Tang et al.~\cite{tang2025revisiting} employ a unified backbone for simultaneous feature extraction and correlation. However, these methods are primarily confined to 2D tracking. In contrast, \textbf{PoseStreamer} extends beyond 2D fusion, utilizing event data to provide robust geometric priors that drive explicit 6DoF pose estimation in 3D space.

\section{Methods}
The overall architecture of PoseStreamer is illustrated in Fig.~\ref{fig:main}. By leveraging the complementary strengths of RGB and event cameras, our framework achieves robust 6DoF pose estimation for unseen moving objects. It consists of three core components:
First, we introduce the \textbf{Adaptive Pose Memory Queue (AMQ)} in Section~\ref{sec:AMQ}, which utilizes historical cues to maintain temporal orientation consistency. 
Next, to address high-speed motion, we propose the \textbf{Multi-modality 3D tracker (M3D)} in Section~\ref{sec:Centers}. This module employs a stereo configuration to provide reliable 3D center priors. 
Finally, the \textbf{Ray Pose Filter (RPF)}, detailed in Section~\ref{sec:RPF}, refines the estimation by sampling hypotheses along the camera ray and selecting the optimal pose via a render-and-compare strategy.


\subsection{RGB-based Instance Pose Initialization}
\label{sec:AMQ}
The primary challenge in tracking moving objects is maintaining orientation consistency across consecutive frames. To address this, we introduce the Adaptive Pose Memory Queue (AMQ), a module designed to stabilize pose estimation by leveraging historical temporal cues.
Initially, to handle unseen objects, we capture surround-view images with a standard RGB camera and reconstruct a reference CAD model using BundleSDF~\cite{wen2023bundlesdf}. An initial pose estimator is then employed to populate a First-In-First-Out (FIFO) queue $M$ of length $N$.
During the tracking phase, AMQ ensures smoothness through an adaptive update strategy. Specifically, at each iteration, historical poses stored in the queue are projected into the Euler-angle space via the mapping $\mathcal{E}^{-1}$. We then apply a confidence scaler $\alpha$ to assign decaying weights to these poses, effectively balancing the contribution of historical trends against current observations. The fused orientation is subsequently mapped back to the rotation matrix domain via $\mathcal{E}$.
By propagating temporal information through low-dimensional pose parameters rather than high-dimensional feature maps $F$, AMQ significantly improves orientation stability with negligible computational overhead. The detailed procedure is outlined in Algorithm~\ref{arg:pose_queue}.

\begin{algorithm}[!ht]
\caption{Adaptive Pose-Queue Update Strategy}
\label{arg:pose_queue}
\KwIn{queue $M$; rotation $R$; 3D centers $C$\;}
\KwOut{pivot rotation $\hat{R}$\;}
$\mathcal{E}$: Euler-to-rotation mapping\;
$\mathcal{E}^{-1}$: rotation-to-Euler mapping\;
$\alpha$: decay weight\;
$\mathcal{H}$: center-pose hypothesis\;
\textcolor[HTML]{006400}{\/// initialization at the first frame.} \\
\If{$i == 0$}{
   $ \hat{R}' \leftarrow \mathcal{H}(C)$\;
}
$N \leftarrow min(i, N) $\;
\textcolor[HTML]{006400}{\/// adaptive scaling orientation} \\
\For{$n = 0$ \textbf{to} $N-1$}{
    $R_n' \leftarrow \mathcal{E}^{-1}(R_{n})$\;
    $a \leftarrow a^{n+1}$\;
    $\hat{R}' \leftarrow \alpha\, \hat{R}' + (1-\alpha)\, R_n'$\;
}
\textcolor[HTML]{006400}{\/// update pose } \\
$\hat{R} \leftarrow \mathcal{E}(\hat{R}')$\;
\KwRet{$\hat{R}$}
\end{algorithm}

\subsection{Multi-modal 3D Center Tracking}
\label{sec:Centers}
\begin{figure}[!ht]
    \centering
\includegraphics[width=0.5\textwidth]{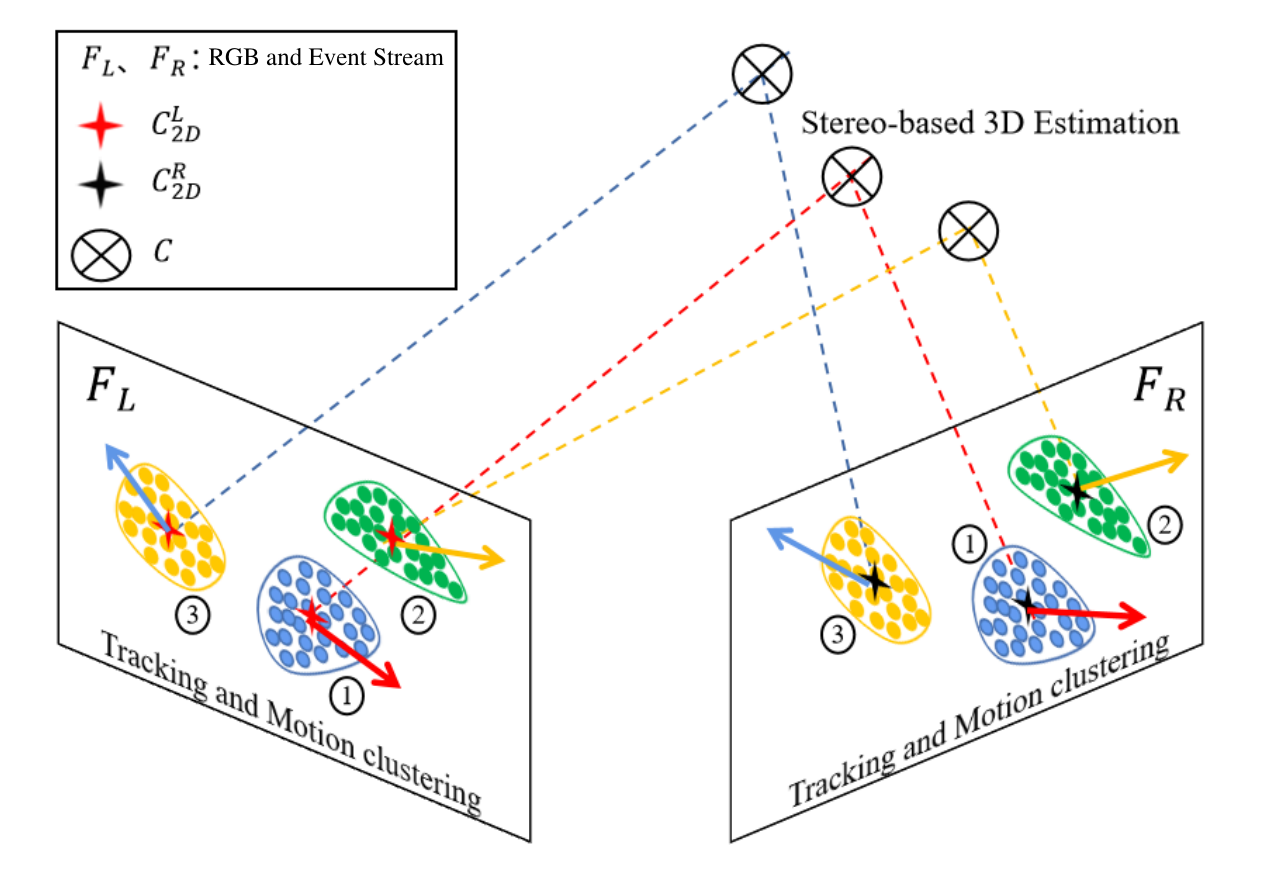}
    \caption{Details of Multi-modality 3D tracker. Features $F_L$ and $F_R$ are extracted from the left and right cameras, respectively, with modalities including RGB images and event stream. Feature points are clustered into groups according to motion consistency, after which the 2D centroids $C_{2D}^L$ and $C_{2D}^R$ are computed from the left and right views. Finally, the 3D object center $\mathbf{C}$ is obtained via disparity-based stereo triangulation.}
    \label{fig:mm3d.drawio}
\label{tab:SDTD}
\end{figure}

Conventional pose estimators relying solely on RGB frames often falter in high-speed scenarios due to severe motion blur and large inter-frame displacements. To overcome this limitation, we leverage complementary modalities: RGB images, which captures high-frequency motion changes, and Event stream, which highlights structural edges. Integrating these inputs, we propose \textbf{M3D}, a lightweight and modality-agnostic tracker. As illustrated in Fig.~\ref{tab:SDTD}, the pipeline extracts features from these stereo streams (RGB and event steam), clusters them based on motion consistency, and robustly computes the 3D center $\mathbf{C}$ via triangulation.

The process begins by extracting synchronized features $F_L$ and $F_R$ from left and right cameras. We initialize a set of feature points $\mathcal{P} = \{ \mathbf{p}_i^{t} \}$ on the image plane and employ Pyramidal Lucas--Kanade optical flow~\cite{lucas1981iterative} to handle rapid motion. This coarse-to-fine approach computes the displacement vector $\Delta \mathbf{p}_i = \mathbf{p}_i^{t+1} - \mathbf{p}_i^{t}$ for each point across consecutive frames.

To distinguish the moving object from the background, we calculate the consistency $C_{ij}$ between points $i$ and $j$ by combining motion displacement and spatial proximity:
\begin{equation}
\resizebox{.88\linewidth}{!}{$
C_{ij} =
\begin{cases}
1, & \big\|\,z([\Delta \mathbf{p}_i,\;\lambda \hat{\mathbf{p}}_i]) -
z([\Delta \mathbf{p}_j,\;\lambda \hat{\mathbf{p}}_j])\,\big\|_2 \leq \tau, \\
0, & \text{otherwise},
\end{cases}
$}
\label{eq:consistency}
\end{equation}

where $z(\cdot)$ denotes z-score normalization, and $\hat{\mathbf{p}} = \mathbf{p}^t / (W,H)$ represents the normalized position. The factor $\lambda$ (set to 0.3) balances the spatial term to ensure clusters are compact.

Finally, we identify the dominant cluster $S_k$ as the target object. We compute its 2D centroids $C_{2D}^{L}=(u_L, v_L)$ and $C_{2D}^{R}=(u_R, v_R)$ from the left and right views, respectively. Defining the stereo disparity as $d = u_L - u_R$, the 3D object center $\mathbf{C}$ is derived via stereo triangulation:
\begin{equation}
\mathbf{C} =
\begin{bmatrix}
X \\ Y \\ Z
\end{bmatrix}
= \frac{b}{d} \cdot
\begin{bmatrix}
u_{L} - c_x^L \\
v_{L} - c_y^L \\
f_x^L
\end{bmatrix},
\label{eq:triangulation}
\end{equation}
where $b$ is the stereo baseline, and $(c_x^L, c_y^L, f_x^L)$ are the intrinsic parameters of the left camera.



\subsection{Ray Pose Filter}
\label{sec:RPF}
To refine the coarse pose estimates provided by M3D and AMQ, we introduce the Ray Pose Filter (RPF). The module operates in three stages: constructing a camera ray from the estimated center, generating pose hypotheses along this ray, and selecting the optimal candidate via an attention-based refinement decoder.

\subsubsection{\textbf{Camera Ray Construction}}
Given the initial 3D object center $\mathbf{C} = [x, y, z]^T$ and rotation $\hat{R}$ from previous stages, we first define the object-centric ray. This ray originates from the camera's optical center and passes through the object's projected center on the image plane.
The 2D projection $(u, v)$ and the depth $d$ of the object center are obtained via the perspective projection:
\begin{equation}
    [u \cdot d, \; v \cdot d, \; d]^T = K \mathbf{C},
\end{equation}
where $K$ is the camera intrinsic matrix. The ray is thus defined as the vector direction passing through pixel $(u, v)$ in the camera coordinate system. This geometric constraint allows us to reduce the search space from a 3D volume to a 1D manifold (the ray), significantly enhancing efficiency.

\subsubsection{\textbf{Ray Pose Generation}}
To handle depth uncertainty, we generate a set of candidate poses along the established ray. We introduce a depth perturbation mechanism to sample hypotheses around the initial depth estimate $d$. The perturbed depth $\hat{d}$ is defined as:
\begin{equation}
    \hat{d} = d + \beta \cdot \mathcal{U}(-1, 1),
\end{equation}
where $\mathcal{U}(-1, 1)$ denotes a uniform distribution, and $\beta$ is a scale parameter set to the mean object diameter. Using these sampled depths, we back-project the 2D center $(u, v)$ to obtain a set of 3D candidate centers $\{\hat{\mathbf{C}}_j\}$. For the $j$-th candidate, the 3D position is reconstructed as:
\begin{equation}
    \hat{\mathbf{C}}_j = \hat{d}_j \cdot K^{-1} [u, \; v, \; 1]^T.
\end{equation}
Combining these positions with the pivotal rotation $\hat{R}$, we form a batch of ray poses. A CUDA-accelerated renderer then generates synthetic feature maps $F_q$ for each candidate pose, which serve as queries for the subsequent refinement.

\subsubsection{\textbf{Pose Refinement}} 
This module functions as both a pose selector and a refiner. In the attention mechanism\cite{amini2021t6d}, the synthetic features $F_q$ (derived from ray poses) serve as the query $q$, while the observed image features $F$ serve as both the key $k$ and value $v$. Formally, we first project the features and then aggregate scale- and perspective-aware context via Multi-Head Attention (MHA):

\begin{equation}
    \begin{aligned}
        q &= \operatorname{FFN}(F_q + \operatorname{Norm}(\operatorname{FFN}(F_q))), \\
        \hat{q} &= \operatorname{MHA}(\operatorname{PE}(q), \; \operatorname{PE}(\operatorname{FFN}(F)), \; \operatorname{FFN}(F)),
    \end{aligned}
\end{equation}
where $\operatorname{FFN}$, $\operatorname{Norm}$, and $\operatorname{PE}$ denote the Feed-Forward Network, Layer Normalization, and Positional Encoding function, respectively. Note that we substitute $k$ and $v$ with the observation feature $F$.

Next, we evaluate the quality of each hypothesis to identify the optimal candidate. We compute a salience score vector $S \in \mathbb{R}^{N \times 1}$ (where $N=64$) for the updated queries $\hat{q}$ and select the best match:
\begin{equation}
    \begin{aligned}
        S &= \operatorname{Softmax}( \operatorname{FFN}(\hat{q}) ), \\
        \hat{q}_{1} &= \operatorname{Topk}(\hat{q},\; S,\;1), \\
        \hat{C}_{1} &= \operatorname{Topk}(\hat{C},\; S,\;1), \\
        \hat{R}_{1} &= \operatorname{Topk}(\hat{R},\; S,\;1), 
    \end{aligned}
\end{equation}
where $\operatorname{Topk}$ selects the candidate with the highest score. 

Finally, two separate FFN heads utilize the selected feature $\hat{q}_{1}$ to predict the residual translation $\Delta C$ and rotation $\Delta R$. The final object center and orientation are updated as follows:
\begin{equation}
\begin{aligned}
 \Delta{C} &= \operatorname{FFN}(\hat{q}_1), &
 \tilde{C} &= \hat{C}_1 + \Delta{C}, \\
 \Delta{R} &= \operatorname{FFN}(\hat{q}_1), &
 \tilde{R} &= \Delta R \cdot \hat{R}_1,
\end{aligned}
\end{equation}
where $\tilde{C}$ and $\tilde{R}$ are the final outputs enqueued into the AMQ for the next iteration. Thanks to this data-driven design\cite{cai2024ov9d}, the refinement decoder is universally applicable to standard RGB and event stream inputs.

\section{Experiments}
\subsection{Dataset}
\subsubsection{MoCapCube6D}
We present \textbf{MoCapCube6D}, a high-precision benchmark for high-speed tracking featuring a Rubik's Cube with synchronized RGB and event streams. Ground truth is obtained via a calibrated MoCap system, ensuring sub-millimeter translation and sub-degree rotation accuracy. The dataset includes three scenes with distinct motion patterns (details in Tab.~\ref{tab:scenes}) and is further partitioned by projected pixel velocity $v$ into \textit{Regular} ($v < 45$ px/s), \textit{Medium} ($45 \leq v < 180$), and \textit{Faster} ($v \geq 180$) to evaluate robustness under diverse dynamics.

\begin{table}[!ht]
\centering
\scriptsize
\caption{Scene definitions used in the benchmark.}
\resizebox{1.0\linewidth}{!}{
\begin{tabular}{p{0.07\linewidth}|p{0.88\linewidth}}
\toprule
\textbf{Scene} & \textbf{Description} \\
\midrule
a & The cube is suspended and rotated rapidly around different axes and at varying angular velocities. \\
\midrule
b & The cube undergoes periodic oscillations resembling pendulum motion, enabling evaluation of phase and frequency stability. \\
\midrule
c & Appearance degradation is introduced through perturbations such as random noise, artificial occlusion, and pixel replacement. \\
\bottomrule
\end{tabular}
}
\label{tab:scenes}
\end{table}

\subsubsection{YCB Object Set}
We also adopt the standard YCB object set\cite{calli2015ycb}. Due to the lack of high-speed multi-modal benchmarks, we collected and labeled real-world sequences of YCB objects using the aforementioned MoCap system. This enables comprehensive qualitative and quantitative validation of our template-free framework on unseen objects.

\subsection{Evaluation Metrics}
To evaluate performance, we employ the standard \textbf{ADD} metric for general objects and \textbf{ADD-S} for symmetric ones like the cube. We report the average recall rate where the mean distance is within 10\% of the object's diameter (ADD(S)@0.1d). Additionally, we provide the mean translation error $\mathbf{e_p}$ (in cm) and rotation error $\mathbf{e_r}$ (in degrees) to offer a more granular analysis.

\begin{table*}[t]
\centering
\scriptsize

\caption{Pose estimation performance under different speed conditions, i.e., Regular, Medium, and Faster, defined by pixel-per-second velocity $v$. Specifically, \textbf{Regular}: $v < 45$, \textbf{Medium}: $45 \leq v < 180$, and \textbf{Faster}: $v \geq 180$. In addition, $e_p(\sigma)$ denotes the translation error (cm), and $e_r(\sigma)$ denotes the rotation error (deg), reported for further evaluation. }
\resizebox{1.0\linewidth}{!}{
\begin{tabular}{c|c|ccc|ccc|ccc}
\toprule
\multirow{2}{*}{\textbf{Scenes}} & \multirow{2}{*}{\textbf{Methods}} & 
\multicolumn{3}{c|}{\textbf{Regular} ($v < 45\,\text{pps}$)} &
\multicolumn{3}{c|}{\textbf{Medium} ($45 \!\leq\! v < 180\,\text{pps}$)} & 
\multicolumn{3}{c}{\textbf{Faster} ($v \geq 180\,\text{pps}$)} \\
&
& ADD(S) & $e_p(\sigma)$ (cm) & $e_r(\sigma)$ (deg)  
& ADD(S)  & $e_p(\sigma)$ (cm) & $e_r(\sigma)$ (deg) 
& ADD(S)  & $e_p(\sigma)$ (cm) & $e_r(\sigma)$ (deg)  \\
\midrule

\multirow{2}{*}{a}
& FoundationPose       & 75.8 & 15.4 & 26.4 & 62.3 & 24.5 & 44.2 & 41.7 & 43.4 & 65.1 \\
& \textbf{Ours}        & \textbf{89.2} & \textbf{8.7}  & \textbf{15.6} & \textbf{78.5} & \textbf{13.2} & \textbf{24.6} & \textbf{65.9} & \textbf{22.5} & \textbf{30.4} \\
\midrule

\multirow{2}{*}{b}
& FoundationPose       & 72.5 & 16.8 & 28.7 & 58.9 & 27.3 & 47.5 & 38.2 & 46.9 & 69.3 \\
& \textbf{Ours}        & \textbf{86.7} & \textbf{9.5}  & \textbf{17.2} & \textbf{75.3} & \textbf{15.1} & \textbf{26.8} & \textbf{62.4} & \textbf{24.8} & \textbf{33.7} \\
\midrule

\multirow{2}{*}{c}
& FoundationPose       & 69.4 & 18.2 & 31.2 & 55.6 & 29.7 & 50.1 & 35.8 & 49.2 & 72.6 \\
& \textbf{Ours}        & \textbf{83.9} & \textbf{10.3} & \textbf{19.5} & \textbf{71.8} & \textbf{16.7} & \textbf{29.4} & \textbf{59.2} & \textbf{26.3} & \textbf{36.2} \\
\bottomrule
\end{tabular}
}
\label{tab:main_result}
\end{table*}

\subsection{Main Results}
\subsubsection{Comparison with State-of-the-Art}
We evaluate PoseStreamer against the state-of-the-art RGB-based method, FoundationPose~\cite{wen2024foundationpose}, on our MoCapCube6D dataset. The results are detailed in Table~\ref{tab:main_result}.
As shown, while FoundationPose performs adequately in slow-motion (\textit{Regular}) scenarios, its accuracy degrades dramatically as object speed increases, suffering from severe motion blur in the \textit{Faster} category. In contrast, PoseStreamer leverages high-temporal-resolution event data to maintain robust tracking across all conditions.

Our method consistently and significantly outperforms FoundationPose. For instance, in the most challenging \textit{Faster} setting for Scene a, PoseStreamer improves the ADD(S) score from \textbf{41.7} to \textbf{65.9} and reduces the rotation error $e_r$ by more than half, from \textbf{$65.1^\circ$} to \textbf{$30.4^\circ$}. This confirms that our multi-modal fusion approach effectively compensates for the deficiencies of RGB-only methods under rapid motion.

\subsubsection{Generalization on Real-world Data}
(This part is logically consistent and can be kept as is.)
We further validate our method on real-world YCB sequences. Despite the lack of ground truth for these ad-hoc recordings, qualitative results demonstrate that PoseStreamer successfully tracks unseen objects without template retraining, handling rapid motions where traditional RGB trackers fail.

\subsection{Ablation Studies}

\subsubsection{Impact of AMQ Queue Size}
In Table~\ref{tab:AMQ_num}, we investigate the influence of the history length $N$ in the Adaptive Pose Memory Queue (AMQ). Increasing $N$ from 0 to 4 yields consistent gains in pose estimation accuracy. Notably, leveraging historical cues ($N \geq 2$) effectively eliminates orientation ambiguity, as evidenced by the \textit{Switch} metric dropping to 0 and the rotation error $e_r$ decreasing to \textbf{$3.7^\circ$}. Consequently, we adopt $N=4$ as the optimal setting, which achieves a peak ADD(S) of \textbf{84.3\%}.

\begin{table}[!ht]
\centering
\caption{Number of frames (N) in AMQ.}
\label{tab:ablation_queue_len}
\scriptsize
\resizebox{0.475\textwidth}{!}{

\begin{tabular}{c|c|c|c|c} 
\toprule
{number frames}& {ADD(S)}$\uparrow$&  $e_r(\sigma)$ (deg) & Switch$\downarrow$&{FPS}$\uparrow$ \\
\toprule
0  & 73.4 & 8.7 & 4  & {45}   \\
1  & 78.3 & 6.2 &  2 & {45}   \\
2  & 81.8 & 4.5 &  0 & {45}   \\
3  & 84.1 & 3.8 &  0 & {45}   \\
\rowcolor[gray]{.9} 
4  & 84.3 & 3.7 &  0 & {45}   \\
\bottomrule
\end{tabular}}
\label{tab:AMQ_num}
\end{table}

\subsubsection{Cumulative Effectiveness}
Table~\ref{tab:ablation} quantitatively details the step-by-step contribution of each module:
\begin{itemize}
    \item \textbf{Baseline vs. +M3D:} The baseline suffers from frequent tracking failures. Integrating the Multi-modality 3D Tracker (M3D) provides robust 2D priors, significantly boosting ADD(S) from \textbf{41.7} to \textbf{67.5} and saturating the 2D projection accuracy (Proj@5pix) to 100\%. However, the high Switch metric \textbf{36.1} indicates that orientation ambiguity remains unresolved.
    \item \textbf{+AMQ:} The addition of AMQ is critical for maintaining temporal orientation consistency. It drastically reduces the Switch metric from \textbf{36.1} to \textbf{3.4} and further improves ADD(S) to \textbf{75.1\%}, demonstrating that historical cues effectively correct pose flipping.
    \item \textbf{+RPF:} Finally, the Ray Pose Filter (RPF) performs fine-grained geometric refinement, culminating in the best performance with a total ADD(S) improvement of \textbf{42.6\%} over the baseline.
\end{itemize}

\begin{table}[!ht]
  \centering
  \scriptsize
  \caption{Cumulative effect: \textcolor[RGB]{0,150,80}{green} values indicate an improvement over the baseline model, while \textcolor{red}{red} values indicate a decrease. Recall of ADD that is less than 0.1 of the object diameter (ADD-S 0.1d).}
  \label{tab:metric}
  \resizebox{0.98\linewidth}{!} {
  \setlength{\tabcolsep}{1.5mm}{
  \begin{tabular}{@{}l|cccc|cc@{}}
    \toprule
    Setting  &ADD$\uparrow$ &ADD(S)$\uparrow$ & 
    Proj@5pix $\uparrow$ &
    Switch $\downarrow$ &
    FPS$\uparrow$  \\
    \midrule
   Baseline  & 3.4 & 41.7 & 8 & 36.4  & 45 \\
   \midrule
   $+$ M3D  & 4.6 & 67.5 & 100 & 36.1  & 45 \\
   $+$ AMQ  & 8.1 & 75.1 & 100 &  3.4  & 45  \\
    $+$ RPF & 10.6 & 84.3 & 100 &  2.8  & 45  \\
   \rowcolor{gray!30} $=$ Totally & 
      \textbf{\textcolor[RGB]{0,150,80}{$+$7.2}} & 
   \textbf{\textcolor[RGB]{0,150,80}{$+$42.6}} & 
    \textbf{\textcolor[RGB]{0,150,80}{$+$92}} & 
   \textbf{\textcolor[RGB]{0,150,80}
   {$-$33.6}} & 
\textbf{{\color{red} $-$0}} \\
   \bottomrule
  \end{tabular}}
  }
  \label{tab:ablation}
\end{table}

\subsubsection{Impact of Ray Pose Sampling Distribution}
\begin{figure}[!h]
    \centering
\includegraphics[width=0.9\linewidth]{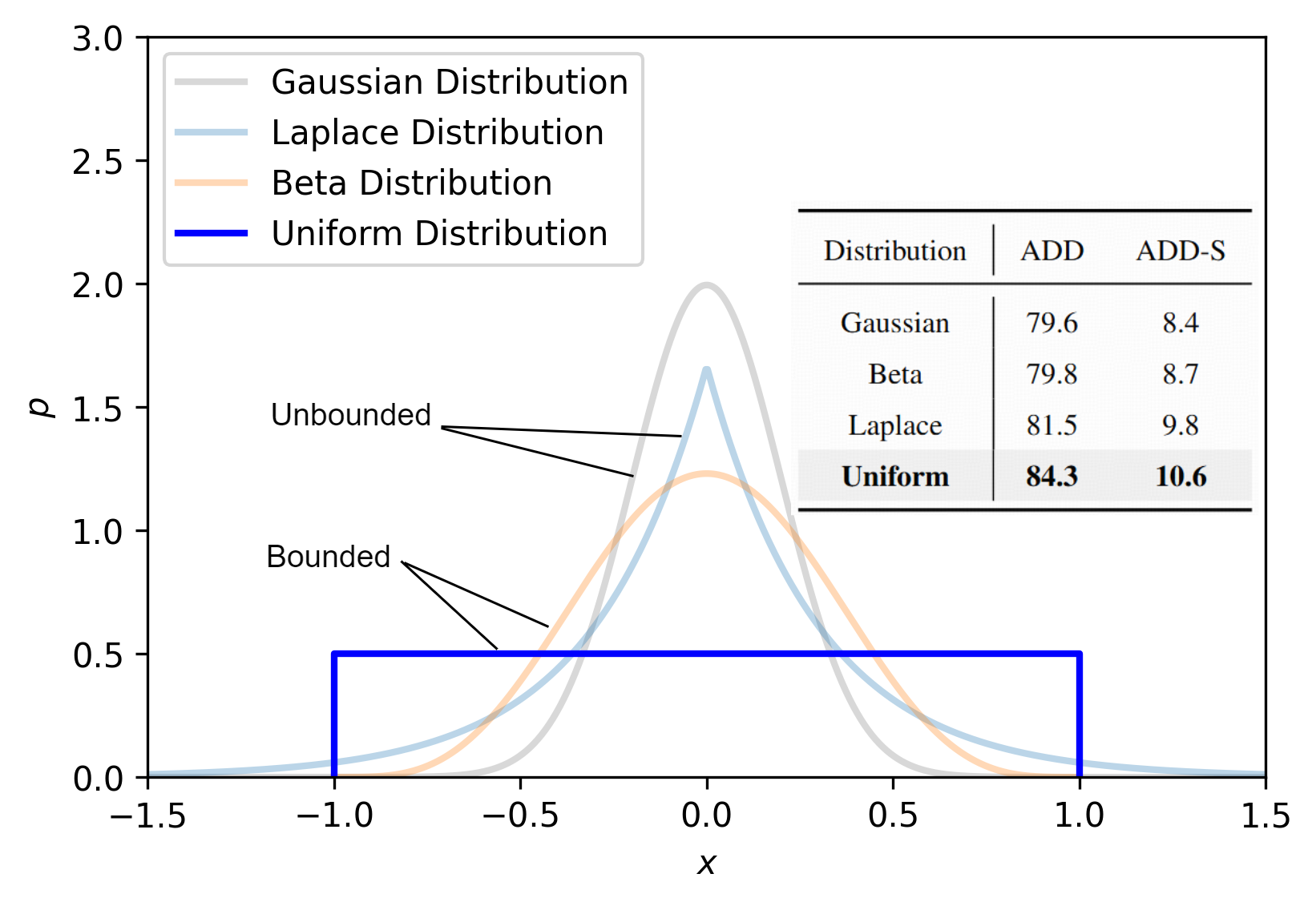}
    \caption{Distribution comparison. showing that the uniform distribution is bounded between -1 and 1, unlike the Laplace and Gaussian distributions, which are unbounded.}
    \label{fig:heatmapCompare}
\label{tab:DC}
\end{figure}

To determine the optimal strategy for the Ray Pose Filter (RPF), we analyze different depth sampling distributions in Fig.~\ref{tab:DC}. We compare the bounded Uniform distribution against unbounded distributions (Gaussian, Laplace, and Beta).
As shown in the embedded table, the Uniform distribution yields the best performance, achieving an ADD of \textbf{84.3\%} and ADD-S of \textbf{10.6}. This is because the Uniform distribution samples candidates within a bounded range defined by the object scale, effectively concentrating hypotheses around the estimated depth. In contrast, unbounded distributions (e.g., Gaussian with \textbf{79.6\%} ADD) tend to sample outliers far from the true depth, introducing geometric noise that degrades refinement accuracy.

\subsection{Runtime Analysis}
Evaluated on an NVIDIA RTX A5000, PoseStreamer requires $\sim$1s for initialization. During tracking, while the M3D module operates at 100 Hz, the rendering-based RPF runs at 45 Hz. Consequently, the overall system achieves \textbf{45 FPS}, maintaining sufficient speed for dynamic scenarios despite the computational overhead.

\label{sec:dis}

\section{Conclusion}
In this paper, we present \textbf{PoseStreamer}, a robust multi-modal framework designed for 6-DoF pose estimation of unseen moving objects in high-speed scenarios. 
To overcome the limitations of standard RGB cameras under rapid motion, we introduce three core components: the Multi-modality 3D Tracker (\textbf{M3D}), which leverages stereo-based multi-modal features to provide reliable 3D center priors. The Adaptive Pose Memory Queue (\textbf{AMQ}), which ensures temporal orientation consistency by utilizing historical cues. And the Ray Pose Filter (\textbf{RPF}), which effectively mitigates depth uncertainty through geometric refinement along camera rays. Furthermore, we construct \textbf{MoCapCube6D}, a high-precision multi-modal benchmark dataset containing synchronized RGB and event streams, to evaluate performance across varying speed profiles. Extensive experiments demonstrate that PoseStreamer significantly outperforms state-of-the-art methods in high-speed settings and exhibits strong generalizability as a template-free framework for unseen objects.


\end{document}